\documentclass[10pt,twocolumn,letterpaper]{article}

\usepackage{iccv}
\usepackage{times}
\usepackage{epsfig}
\usepackage{graphicx}
\usepackage{amsmath}
\usepackage{float}
\usepackage{amssymb}
\usepackage{wrapfig}
\usepackage[inline]{enumitem}
\usepackage[numbers]{natbib}

\usepackage[breaklinks=true,bookmarks=false]{hyperref}

\iccvfinalcopy 

\def\iccvPaperID{43} 

\ificcvfinal\fi

\newcommand{\s}{$\theta_{\text{specialize}}$ }
\newcommand{\R}{$\theta_{\text{reuse}}$ }
\newcommand{\M}{$\mathcal{M}$ }
\newcommand{\W}{$\mathcal{W}$ }

\begin{document}

\title{Differentiable Weight Masks for Domain Transfer}

\author{Samar Khanna$^*$\\
\and
Skanda Vaidyanath$^*$\\
\and 
Akash Velu $^*$\\
\and
{\small Stanford University} \\
{\small \texttt{\{samarkhanna,svaidyan,avelu\}@cs.stanford.edu}} \\
}

\maketitle
\ificcvfinal\thispagestyle{empty}\fi

\def\thefootnote{*}\footnotetext{Equal contribution. Published in Out of Distribution Generalization in Computer Vision (OOD-CV) workshop at ICCV 2023.}
\renewcommand*{\thefootnote}{\arabic{footnote}}
\setcounter{footnote}{0}

\begin{abstract}
One of the major drawbacks of deep learning models for computer vision has been their inability to retain multiple sources of information in a modular fashion.
For instance, given a network that has been trained on a source task, we would like to re-train this network on a similar, yet different, target task while maintaining its performance on the source task. 
Simultaneously, researchers have extensively studied modularization of network weights to localize and identify the set of weights culpable for eliciting the observed performance on a given task.
One set of works studies the modularization induced in the weights of a neural network by learning and analysing weight masks.
In this work, we combine these fields to study three such weight masking methods and analyse their ability to mitigate ``forgetting'' on the source task while also allowing for efficient finetuning on the target task.
We find that different masking techniques trade-off retaining knowledge in the source task with learning to perform well on the target task.
\end{abstract}

\section{Introduction}

Deep learning algorithms have proven to be extremely successful in  a wide range of supervised learning tasks \cite{https://doi.org/10.48550/arxiv.2005.14165} \cite{https://doi.org/10.48550/arxiv.2204.06125} in vision and language.
These methods are capable of utilizing large datasets to learn models which can generalize to new datapoints which are within the training data distribution. Although some particularly large models such as GPT-3 and DALLE-2 have exhibited emergent qualities of generalizing to \emph{new} data distributions, smaller models in vision and language tend to suffer from \emph{distribution shift}.

Often, when practitioners want to improve the performance of their model on a new domain that is out of distribution from the one they trained on, a natural idea is to fine-tune the weights of the model with data from this new domain.
This often tends to improve the performance on the new (target) domain but tends to lose performance on the original (source) domain.
However, we'd like models to be able to maintain their performance on the source domain while improving on the target domain. For instance, a model used in a self-driving car which is adapted to a new driving environment should retain strong performance in the original environment it was trained on, while achieving competitive performance in the new environment.
Another example is when an image classifier trained on paintings is fine-tuned on a new domain of cartoon images, and it loses its performance on the original painting domain.
This is related to the \emph{catastrophic forgetting} problem in continual learning where a model trained on a new task tends to forget older tasks it learned.

In this work, we examine methods by which models can be adapted to a new domain \emph{without} experiencing significant forgetting. 
One way to achieve this could be by modularizing the network and splitting the weights as being specific to the source domain and others that can be edited to improve performance on the target domain without a drop in the source domain.
We take inspiration from methods that edit the weights of a trained model as a means of adapting its performance.
Specifically, we focus on differentiable weight masks \cite{csordas} and model editor networks literature such as \cite{mitchell2022fast}, methods which either directly modify or mask the weight matrices of a trained neural network in order to analyze the network's properties or to edit its performance on specific datapoints. 
In our setting, we use methods similar to these to identify weights that can be modified to retain source domain performance while achieving improved target domain performance. 

We analyse these different masking methods on an image classification task and present that trade-offs that each method offers in this paradigm.
\section{Related Work}
Our setting is similar to transfer learning, where a model trained on one dataset is finetuned on another dataset \cite{5206848, Devlin2019BERTPO, Brown2020LanguageMA} and domain adaptation, where the learning algorithm has access to training data from \emph{multiple} source domains and must transfer to a target domain \cite{10.5555/2946645.2946704, Guo2018MultiSourceDA}.
However, we differ from both because our learning algorithm must adapt to a distribution shift (typically in the inputs) and not a new prediction task and we explicitly care about performance in the first task in addition to the model's performance on the transfer task. Further, we assume access to only one source domain, in contrast to several domain adaptation methods.

Multi-task learning \cite{rosenbaum2018routing, yang2020multitask} is another setting that tackles the problem of simultaneously learning multiple learning tasks at once. However, in this setting, methods typically assume simultaneous access to all training tasks of interest.
Our setting perhaps is closest to a two-task continual learning paradigm \cite{doi:10.1073/pnas.1803839115, 10.5555/3305890.3306093} where tasks are presented in a sequential manner and at the end of training, the model is evaluated on \emph{all tasks}, and must hence retain predictive performance on previous tasks.

We also note that prior work \cite{luo2017entropybased, Lin_2019_CVPR, 8416559} has tackled the problem of pruning the weights and filters of a CNN but this was done with the goal of network optimization rather than domain generalization.

\section{Problem Setup}
Consider a source domain task $\mathcal{S}$ and a model trained on this domain, $f_\theta$, using a source dataset $\mathcal{D_S}$. 
\emph{We assume that we no longer have access to $\mathcal{D_S}$ once the model has been trained on it.}
We have a target domain task $\mathcal{T}$ we would like to generalize to.
We would like to modify $f_\theta$ to obtain a model $f_{\theta '}$ that performs well on both the source and target domains.
This modification must be achieved with a dataset $\mathcal{D_T}$ from the target domain. 
We would like to achieve this by modularizing the weights of $f_\theta$ by splitting the parameters $\theta$ into \s{} and \R{} as described in~\cite{csordas}.

\section{Method}

The core idea is that modularizing the weights of a pretrained network into \R and \s can mitigate forgetting on $\mathcal{S}$.
Concretely, we can freeze the \s weights when fine-tuning on $\mathcal{T}$ so we don't change the weights that are important to $\mathcal{S}$.

There are several methods one can think of to split the weights into \s and \R. One common theme is to generate a decision function that determines whether each individual weight $w$ can be modified or not. 
For a given layer of weights $\mathcal{W} \in \mathbb{R}^D$, we would like to learn a binary mask $\mathcal{M} \in \{0, 1\}^D$, where each weight can be changed during finetuning if its mask value is 0, and vice-versa.

\subsection{Naive masking}
\label{sec:naive}

\begin{wrapfigure}{l}{0.25\textwidth}
    \centering
    \includegraphics[width=\linewidth]{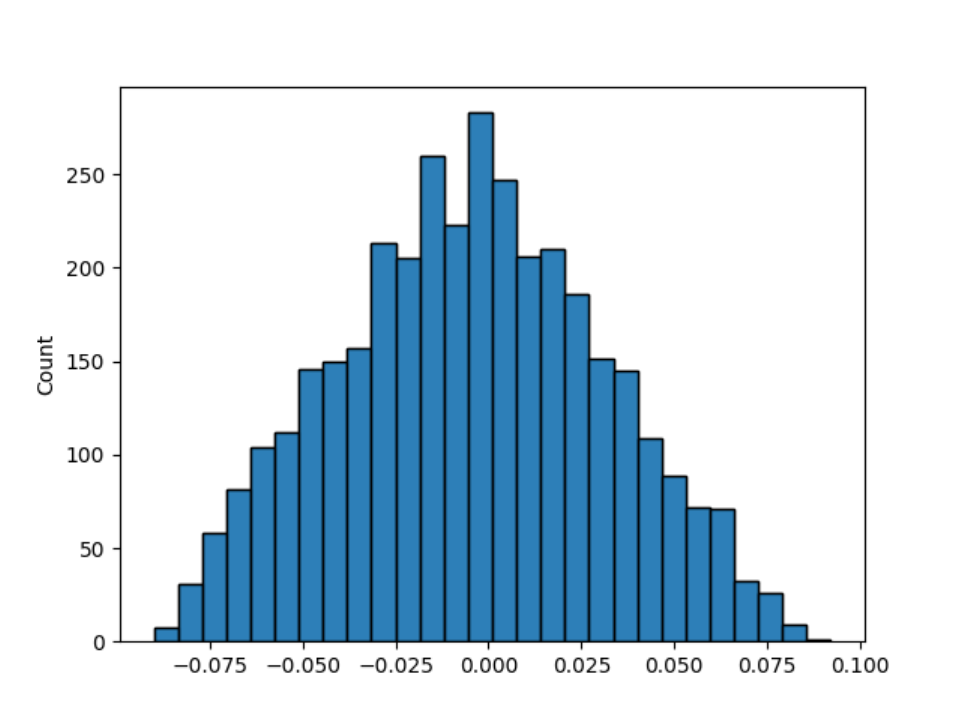}
    \caption{Weight distribution after training on the Photo domain}
    \vspace{-5pt}
    \label{fig:dist}
\end{wrapfigure}

First, we try a naive method to learn a mask \M{} from the weights \W{}. 
To this end, we first analyze the distribution of the weights in \W{} and note that the trained weights are normally distributed (fig. \ref{fig:dist}).

One way to interpret this distribution is that weights that are further away from the mean are more important for the task than weights that are close to the mean. 
The idea is that the ``extreme'' weights are important to the source task and represent \s{}. These must be kept frozen for the fine-tuning process.

Concretely, let the mean of the weights in \W{} be $\mu$ and the standard deviation be $\sigma$. 
We set \M{} such that the values in \M{} are 1 if $(w > \mu + \sigma) \mid (w < \mu - \sigma)$ where $w$ is each weight in \W{}.

\subsection{Editor networks based}
\label{sec:edit}
Directly learning a mask \M{}, while ideal, can prove to be challenging since discrete distributions cannot be differentiated through easily \cite{gumbel, oord2018neural}. 
One solution is to borrow concepts from model editor networks literature \cite{mitchell2022fast, DeCao2021EditingFK} to learn an auxiliary $\Delta$\W{} $\in \mathbb{R}^D$ from which we can recover \M{} using simple thresholding.
We train a model on some source task $\mathcal{S}$ to convergence and let the last layer weights of this learned model be \W{}. 
We would like to learn a $\Delta$\W{} such that we can edit the weights of the learned model by replacing \W{} with \W{} + $\Delta$\W{} \emph{which we get by training on the same source domain.} 
$\Delta$\W{} should allow us to make as many edits as possible to the already tuned weights \W{} without degrading performance on the source domain.
If it is possible to edit the weight while maintaining performance on the source domain, then the weight probably does not belong to \s{}. 
We would also like \s{} to be a small set so we have enough network capacity when we finetune on the target domain.
Hence, we add an L1 penalty that encourages the  $\Delta$\W{} values to be non-zero.
So the overall objective is to reduce the cross-entropy loss (in the source domain) while also increasing this L1 penalty.
Note that this objective only updates $\Delta$ \W{} and not \W{} itself.

Now that we have our trained auxiliary mask, $\Delta$\W{}, we can recover a mask \M{} by heuristically thresholding the values in $\Delta$\W{}. Following a similar strategy from section~\ref{sec:naive}, we first compute the magnitude of $\Delta$\W{} by taking the absolute value of its entries and freeze weight values more than one standard deviation from the mean.

\subsection{Binary masks}
\label{sec:mask}
We now explore a method that directly learns a binary mask.
If we can learn a \M{} directly, we can simply freeze the weights indicated by a 1 in \M{} and fine-tune the rest. The weights that are getting masked out should not be very useful for the source task since the performance on the source task doesn't degrade when we mask these weights.

In general, since we cannot differentiate through discrete sampling operations, it is difficult to learn discrete masks \M{} with gradient based methods.
However, \citet{gumbel} show that the Gumbel distribution can be used to generate discrete samples in a differentiable fashion and has been used successfully in downstream works \cite{csordas,igor}. We encourage the reader to refer to the original paper \cite{gumbel} for a detailed description of the technique.

The training procedure is as follows: once we have a pre-trained model on the source domain, we would like to learn a binary weight mask \M{} that will replace the final \W{} of the model with \W{} * \M{} where * represents an element-wise multiplication operation. 
Our objective is to decrease cross-entropy loss while masking out as many weights as possible. 
We use a penalty term similar to the one we saw in the previous section to achieve this.
We find that this penalty term is crucial in recovering a fairly sparse mask at the end of training.
During training, for stability, we sample multiple masks per batch as prescribed by~\citet{csordas}.

\subsection{Using the trained masks for fine-tuning}
\label{sec:use_mask}

Once we have trained our masks on the source domain $\mathcal{S}$, we can fine-tune our model on the new target domain $\mathcal{T}$, while not losing performance on the source domain. 
For all parameters $\theta_i$ such that $b_i = 1$ (i.e. $\theta_i \in \theta_{\text{specialize}}$), we let $\theta_i = \theta_i^{\mathcal{S}\ast}$, where $\theta^{\mathcal{S}\ast}$ denotes the fully trained weights on the source domain. For all parameters $\theta_j$ such that $b_j = 0$ (i.e. $\theta_j \in \theta_{\text{reuse}}$), we have three options: 
\begin{enumerate}
  \item[(i)] $\theta_j = \theta_j^{\mathcal{S}\ast} $, where we start updating $\theta_j$ from its final trained value on $\mathcal{S}$
  \item[(ii)] $\theta_i^{\mathcal{S}}$, where $\theta^{\mathcal{S}}$ denotes the original initialisation of the weights for the model \textit{before} it was trained on the source domain $\mathcal{S}$. This motivation is inspired from the Lottery Ticket Hypothesis ~\cite{frankle2018lottery}.
  \item[(iii)] $\theta_i^{\text{random}}$, where we consider a fully random re-initialisation of $\theta_{\text{reuse}}$. 
\end{enumerate} 
We study the effect of these initialization strategies in section \ref{sec:ablations}. Our default initialization strategy is (ii).

\section{Experiments}
\begin{figure*}[h]
    \centering
    \includegraphics[width=0.75\linewidth]{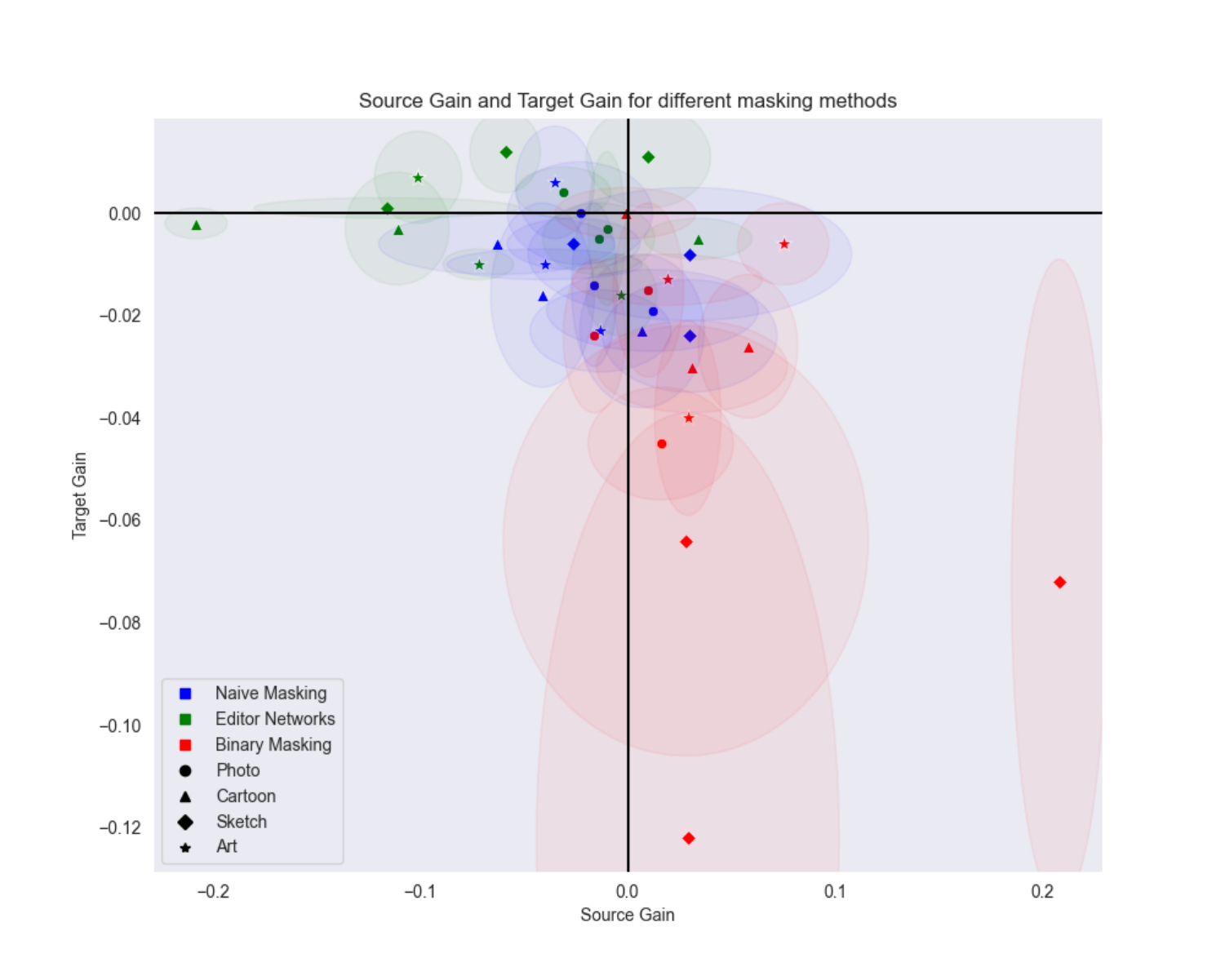}
    \vspace{-10pt}
    \caption{Performance of different masking strategies on the source and target domains after fine-tuning on the target domain. The \textit{x-axis} represents the gain in accuracy in source performance when compared to a model that is finetuned directly on the target domain and the \textit{y-axis} the gain in accuracy on target domain performance under the same setting. 
    Ideally we would like large positive values on both axes.
    The shapes indicate the different source domains as shown in the legend and the colors indicate the different masking strategies. The shading represents the standard deviation on each axis and the center is the mean across 3 seeds. For binary masking, the variance in target gain is larger than for source gain, which still demonstrates its increased capacity to avoid ``forgetting".}
    \label{fig:perf}
\end{figure*}

\begin{table*}[t!]
\centering
\begin{tabular}{|c|cc|cc|cc|cc|}
\hline
Domain       & \multicolumn{2}{c|}{Photo}                             & \multicolumn{2}{c|}{Art}                              & \multicolumn{2}{c|}{Cartoon}                          & \multicolumn{2}{c|}{Sketch}                           \\ \hline
Method & \multicolumn{1}{c|}{$\mathcal{S}$ Gain}  & $\mathcal{T}$ Gain  & \multicolumn{1}{c|}{$\mathcal{S}$ Gain}  & $\mathcal{T}$ Gain  & \multicolumn{1}{c|}{$\mathcal{S}$ Gain}  & $\mathcal{T}$ Gain  & \multicolumn{1}{c|}{$\mathcal{S}$ Gain}  & $\mathcal{T}$ Gain     \\ \hline
$\mathcal{S}$ Start & \multicolumn{1}{c|}{\textbf{0.009}}  & \textbf{-0.028} & \multicolumn{1}{c|}{0.041}          & \textbf{-0.020} & \multicolumn{1}{c|}{\textbf{0.029}} & \textbf{-0.019}          & \multicolumn{1}{c|}{\textbf{0.088}} & -0.086 \\ \hline
$\mathcal{S}$ End   & \multicolumn{1}{c|}{\textbf{0.009}} & -0.029         & \multicolumn{1}{c|}{0.050}          & -0.054         & \multicolumn{1}{c|}{0.026}          & -0.036 & \multicolumn{1}{c|}{0.057}          & -0.050         \\ \hline
Random       & \multicolumn{1}{c|}{0.005} & \textbf{-0.028 }         & \multicolumn{1}{c|}{\textbf{0.051}} & -0.056          & \multicolumn{1}{c|}{0.026}          & -0.036         & \multicolumn{1}{c|}{0.054}          & \textbf{-0.048 }      \\ \hline
\end{tabular}
\label{tbl:ablations}
\vspace{5pt}
\caption{Ablation study to determine best method of initialisation for $\theta_{\text{reuse}}$. Refer to section \ref{sec:metrics_bounds} for a description of $\mathcal{S}$ Gain and $\mathcal{T}$ Gain. 
}
\end{table*}

In this section we explore the effect of varying masking strategies with respect to (i) performance on the target domain and (ii) performance on the source domain after fine-tuning on the target domain.

\textbf{Dataset} We conduct our experiments on the PACS dataset \cite{pacs}.
PACS is an image dataset for domain generalization. 
It consists of four domains, namely Photo (1,670 images), Art Painting (2,048 images), Cartoon (2,344 images) and Sketch (3,929 images). 
Each domain consists of the same seven classification categories: dog, elephant, giraffe, guitar, house, and person.
We wish to study how training on a source domain (eg: Photos), then training on a target domain (eg: Sketch), affects the performance of the model on the original source domain (eg: Photos).

\textbf{Model} We use a ResNet-18 \cite{resnets} pretrained with ImageNet weights \cite{imagenet}. Early experiments determined that using a larger model or finetuning all weights yield negligible improvements in accuracy compared to tuning just the last layer. Therefore, we finetune and mask only the final layer (the MLP head) of the model in all our experiments below.

\subsection{Upper and Lower Bounds}
\label{sec:metrics_bounds}
We compare each masking method relative to finetuning all weights of the MLP head on $\mathcal{D_S}$ (similar to the masking methods), and then also to finetune all weights of the MLP head on $\mathcal{D_T}$. We denote this as unmasked finetuning.
We expect unmasked finetuning to show the largest drop in performance on $\mathcal{S}$ and the biggest gain on $\mathcal{T}$ compared to all masking methods, demonstrating the problem of ``forgetting" by giving us approximate lower and upper bounds, respectively. Thus, rather than plotting the absolute accuracy of each masking method and unmasked finetuning, we compare the difference in accuracy between each masking method and unmasked finetuning on:
\begin{enumerate*}[label=(\roman*)]
  \item the source domain $\mathcal{S}$, denoted as \textbf{source gain}, to indicate the increase in performance achieved by the masking method on the source domain over unmasked finetuning
  \item the target domain $\mathcal{T}$, denoted as \textbf{target gain}, reasoned similarly.
\end{enumerate*}
These two metrics allow us to compare the benefits of each masking method relative to each other, and across domains.



\subsection{How do the different masking strategies compare against each other?}

We plot the source gain and target gain for each masking method in figure \ref{fig:perf}. We find that learned binary masks are much better than other masking methods on the source gain metric, thus ``remembering" more from the source domain even after being finetuned on the target domain. However, this comes at the cost of decreased performance on the target domain. Naive masking shows no discernible improvement or regression on source gain, but is consistently lower than unmasked tuning on the target domain. Finally, editor-based masking shows improvements on the target domain but at the cost of ``forgetting" more on the source domain. Across all masking methods, there seems to be a non-linear tradeoff in improving source domain performance at the cost of decreasing performance on the target domain.
Interestingly, directly learning binary masks does more to combat forgetting than editor-based masks, while the latter un-freezes more of the right weights to improve target gain.

\subsection{How does weight initialization affect the target-domain performance of binary masking?}
\label{sec:ablations}

We compare the three different weight initialization strategies for the binary masking method as described in section \ref{sec:use_mask}. 
The results are shown in table \ref{tbl:ablations}.
The different initialization methods perform similarly but we notice that using initial weights from $\mathcal{S}$ as described in method (ii) in section \ref{sec:use_mask} is almost consistently better and this is in line with the results from \citet{frankle2018lottery}.

\section{Conclusion and Future Work}
In this work, we compared the efficacy of 3 different masking strategies and their effect on alleviating the forgetting problem on transfer learning settings.
We showed that the binary masking strategy is effective at alleviating forgetting while the real-valued masking strategies are better at target domain performance.
We also study the effect of different weight initializations in the fine-tuning process.

For future work, we would like to delve into the reason behind the differences in source and target gain performance across the masking methods. Additionally, we would explore more complex masking strategies that are founded in Information Theory and linear algebra ideas such as low-rank approximations of weights.
We also hope to investigate continual learning and multi-task settings where there are more than two domains to generalize to.


{\small
\bibliographystyle{plainnat}
\bibliography{egbib}
}

\end{document}